\documentclass[]{spie}  %>>> use for US letter paper
\usepackage{amsmath,amsfonts,amssymb}
\usepackage{graphicx}
\usepackage[colorlinks=true, allcolors=blue]{hyperref}
\usepackage{float}
\usepackage[caption = false]{subfig}
\usepackage{siunitx}
\title{Improving utility of brain tumor confocal laser endomicroscopy: objective value assessment and diagnostic frame detection with convolutional neural networks}

\author[a,b]{Mohammadhassan Izadyyazdanabadi}
\author[a,b,c]{Evgenii Belykh}
\author[b]{Nikolay Martirosyan}
\author[b]{Jennifer Eschbacher}
\author[b]{Peter Nakaji}
\author[a]{Yezhou Yang}
\author[b]{Mark C. Preul}
\affil[a]{Arizona State University, Tempe AZ 85281, USA}
\affil[b]{Barrow Neurological Institute, Saint Joseph's Hospital and Medical Center, 350 W Thomas Rd, Phoenix, AZ 85013}
\affil[c]{Irkutsk State Medical University, Krassnogo vosstaniya 1, Irkutsk, Russia 664003}

\authorinfo{Further author information: (Send correspondence to M.I)\\M.I.: E-mail: mizadyya@asu.edu }

\begin{document} 
\maketitle

\begin{abstract}
Confocal laser endomicroscopy (CLE), although capable of obtaining images at cellular resolution during surgery of brain tumors in real time, creates as many non-diagnostic as diagnostic images. Non-useful images are often distorted due to relative motion between probe and brain or blood artifacts. Many images, however, simply lack diagnostic features immediately informative to the physician. Examining all the hundreds or thousands of images from a single case to discriminate diagnostic images from nondiagnostic ones can be tedious. Providing a real time “diagnostic value” assessment of images (fast enough to be used during the surgical acquisition process and accurate enough for the pathologist to rely on) to automatically detect diagnostic frames would streamline the analysis of images and filter useful images for the pathologist/surgeon. We sought to automatically classify images as diagnostic or non-diagnostic. AlexNet, a deep-learning architecture, was used in a 4-fold cross validation manner. Our dataset includes 16,795 images (8572 nondiagnostic and 8223 diagnostic) from 74 CLE-aided brain tumor surgery patients. The ground truth for all the images is provided by the pathologist. Average model accuracy on test data was 91\% overall (90.79 \% accuracy, 90.94 \% sensitivity and 90.87 \% specificity). To evaluate the model reliability we also performed receiver operating characteristic (ROC) analysis yielding 0.958 average for area under ROC curve (AUC). These results demonstrate that a deeply trained AlexNet network can achieve a model that reliably and quickly recognizes diagnostic CLE images.
\end{abstract}

\keywords{Confocal laser endomicroscopy, Convolutional neural networks, Computer aided diagnosis, Image quality assessment, Brain tumor surgery, Precision surgery  }

\section{INTRODUCTION}
\label{sec:intro}  % \label{} allows reference to this section

Handheld, portable Confocal Laser Endomicroscopy (CLE) is being explored for neurosurgery of brain tumors because of its ability to image histopathological features of the tissue in real time during surgery. CLE provides for the first time imaging during brain tumor surgery at cellular resolution, and thus a significant technology advancement for precision brain tumor surgery toward surgery on a cellular resolution\cite{charalampaki2015confocal,sanai2011intraoperative}.

\begin{figure}
\centering
\subfloat[]{\includegraphics[width = 1.35in]{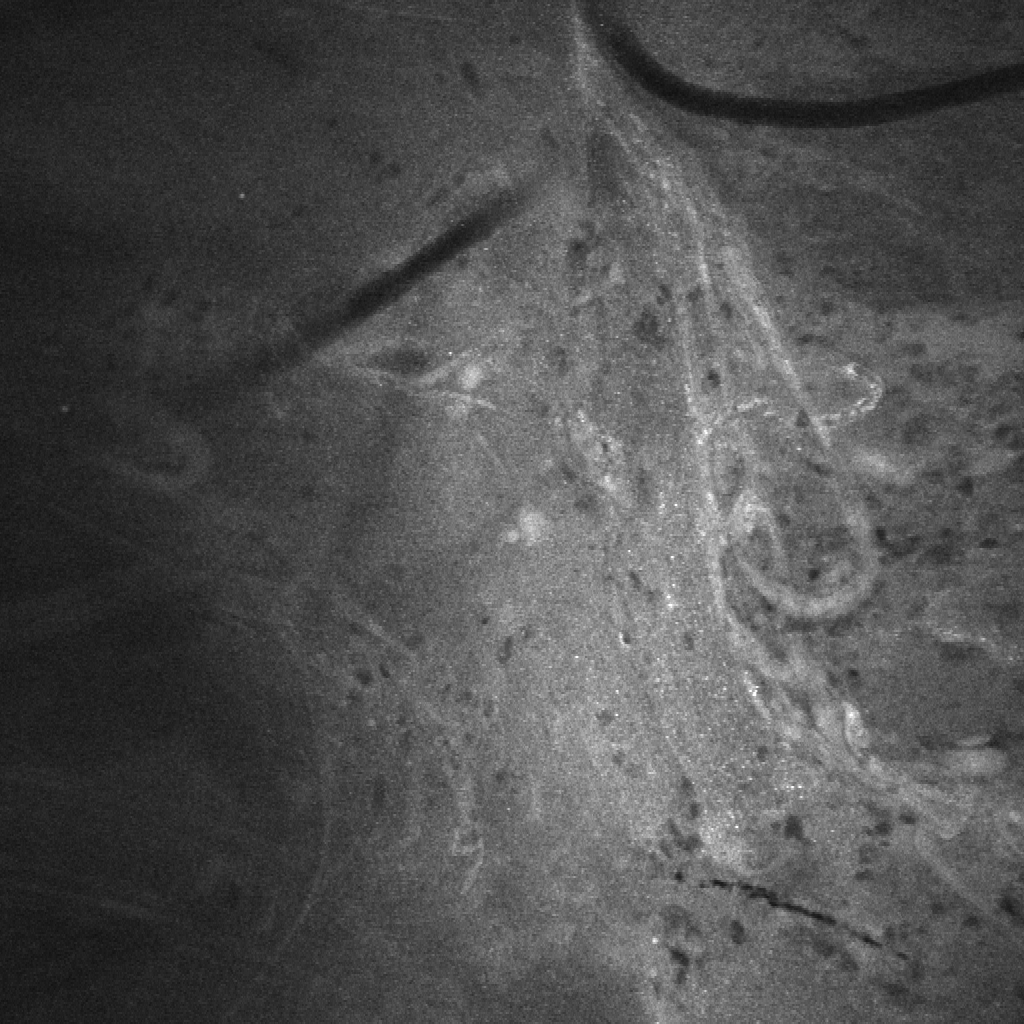}} 
\subfloat[]{\includegraphics[width = 1.35in]{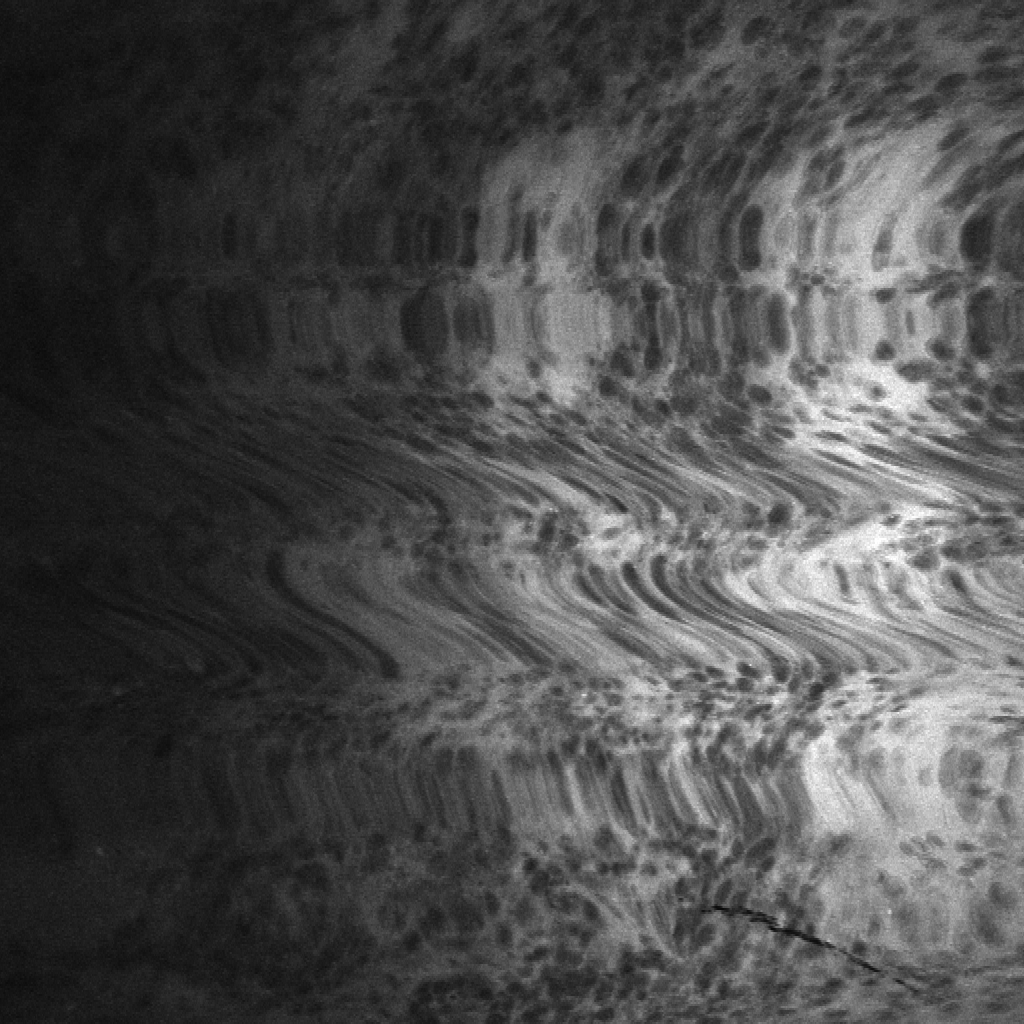}}
\subfloat[]{\includegraphics[width = 1.35in]{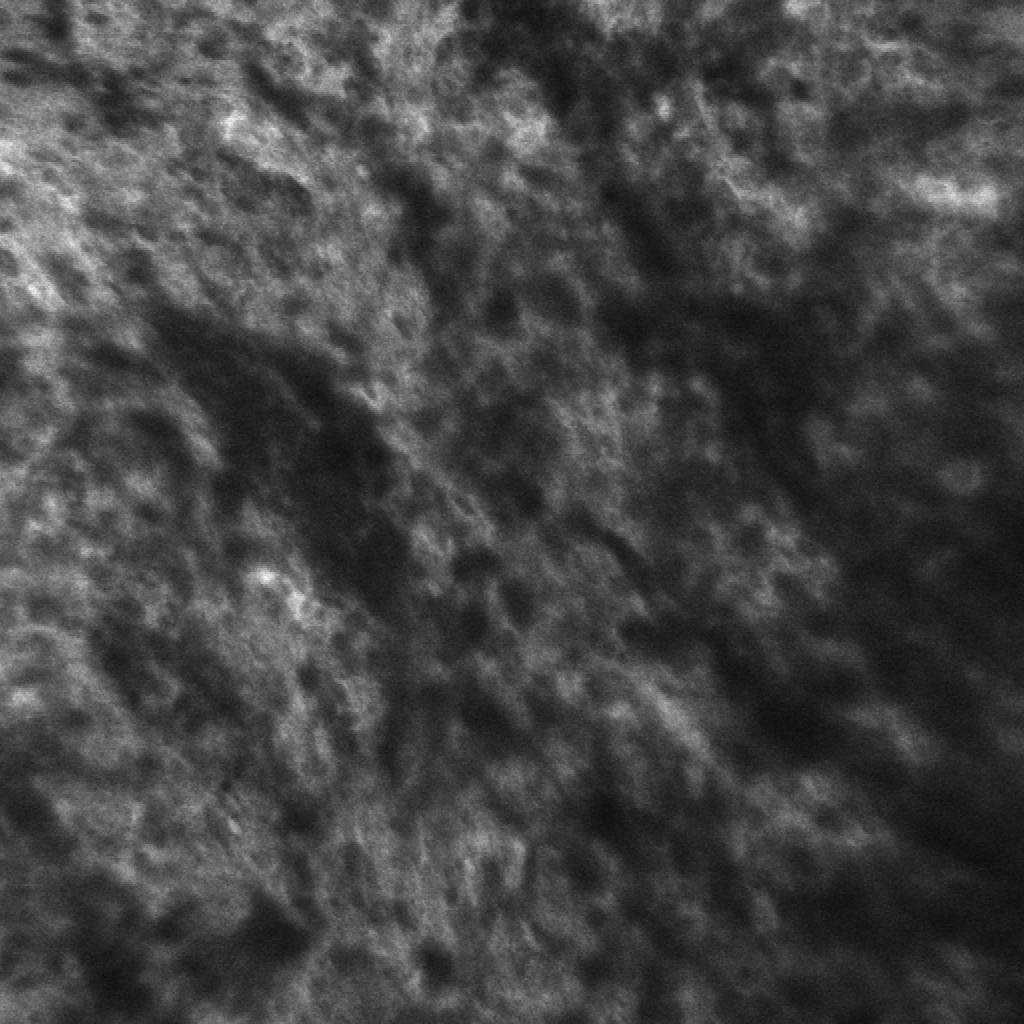}} 
\subfloat[]{\includegraphics[width = 1.35in]{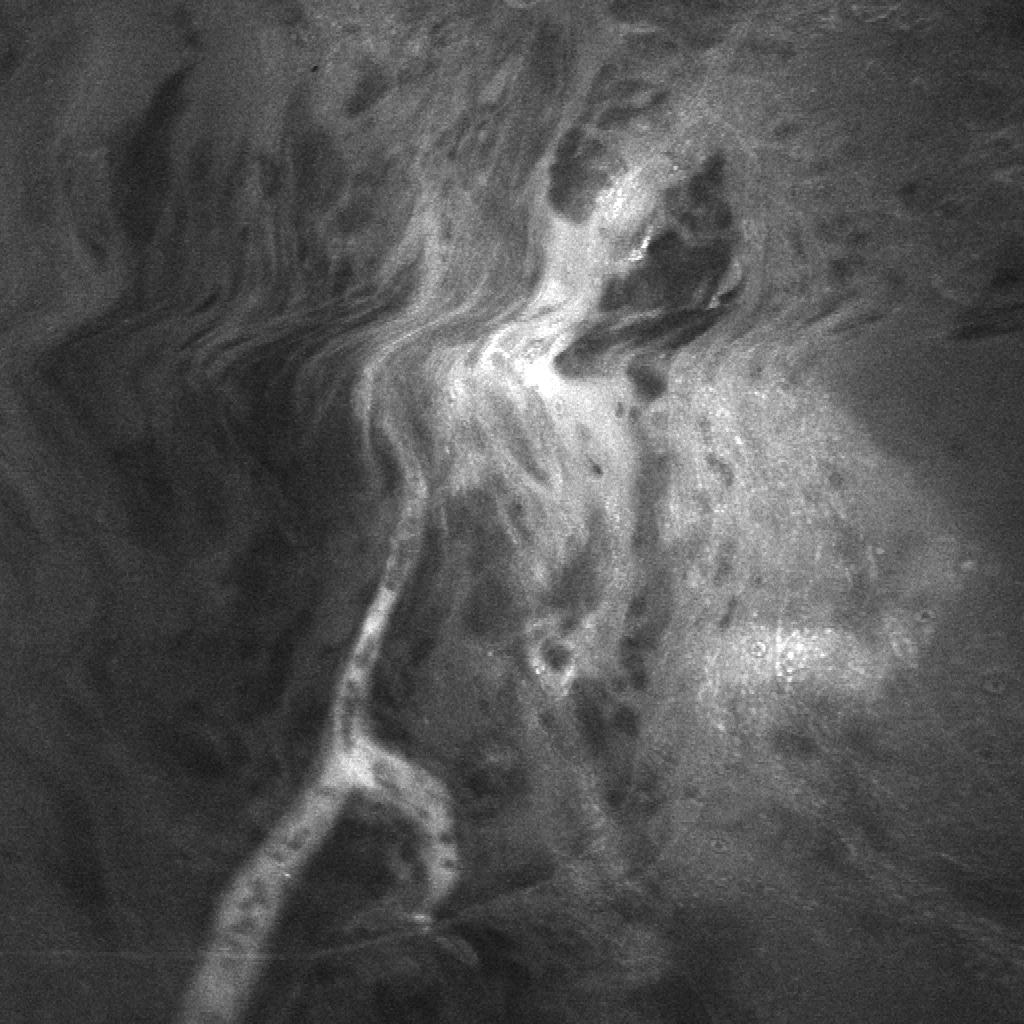}}\
\subfloat[]{\includegraphics[width = 1.35in]{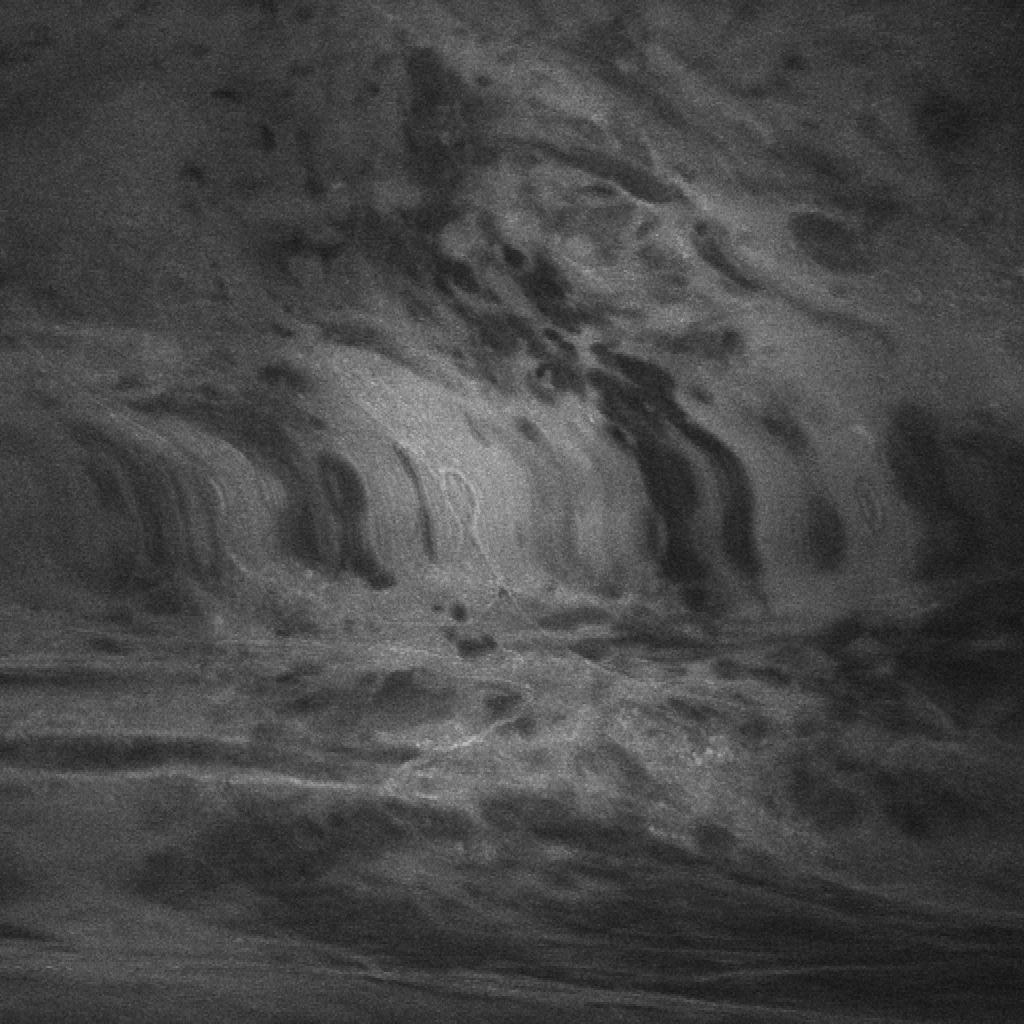}} 
\subfloat[]{\includegraphics[width = 1.35in]{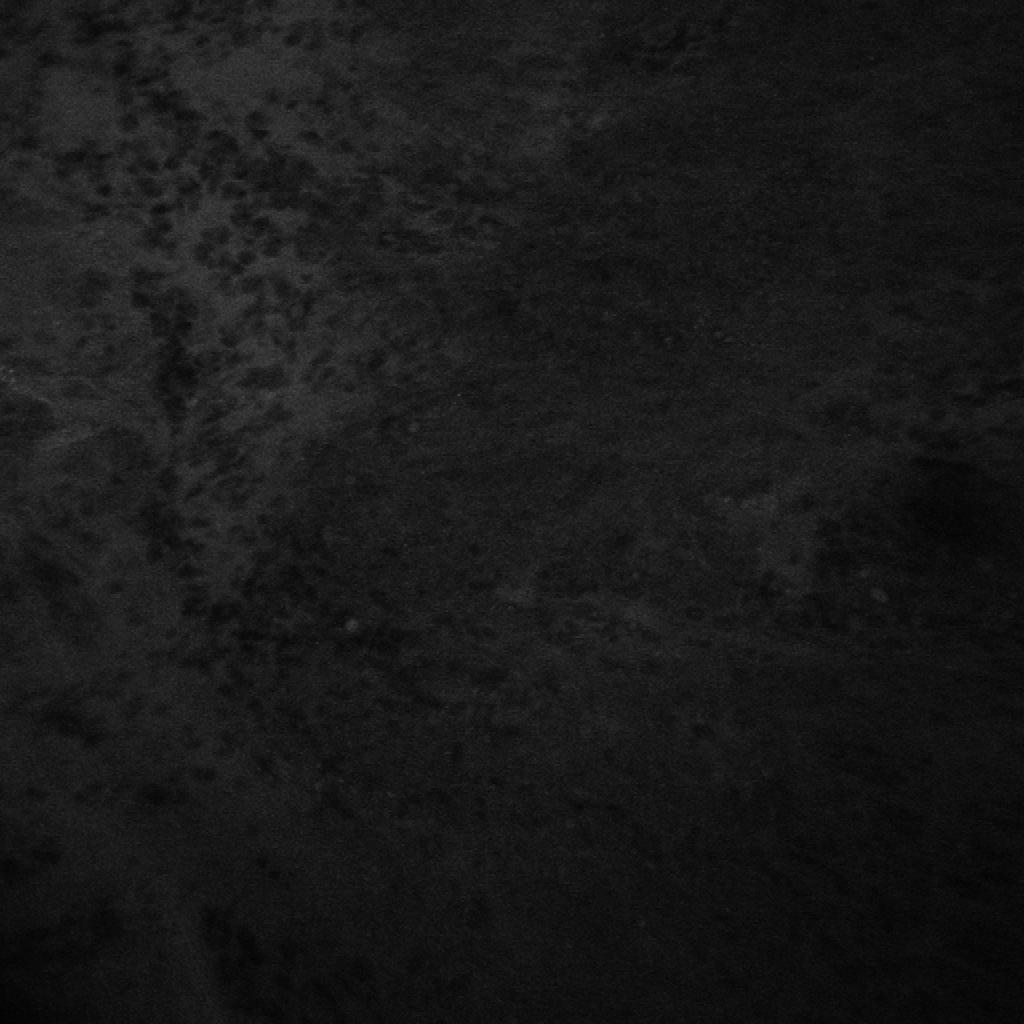}} 
\subfloat[]{\includegraphics[width = 1.35in]{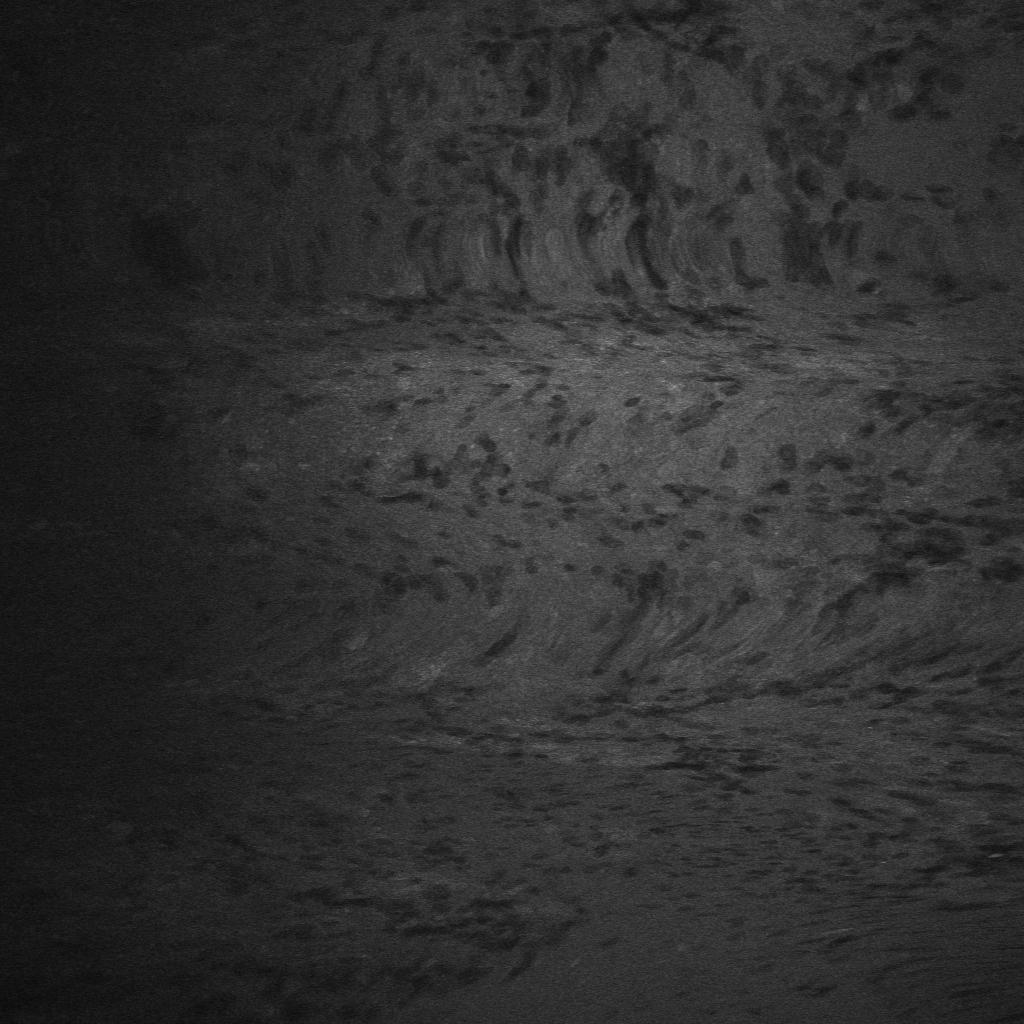}}
\subfloat[]{\includegraphics[width = 1.35in]{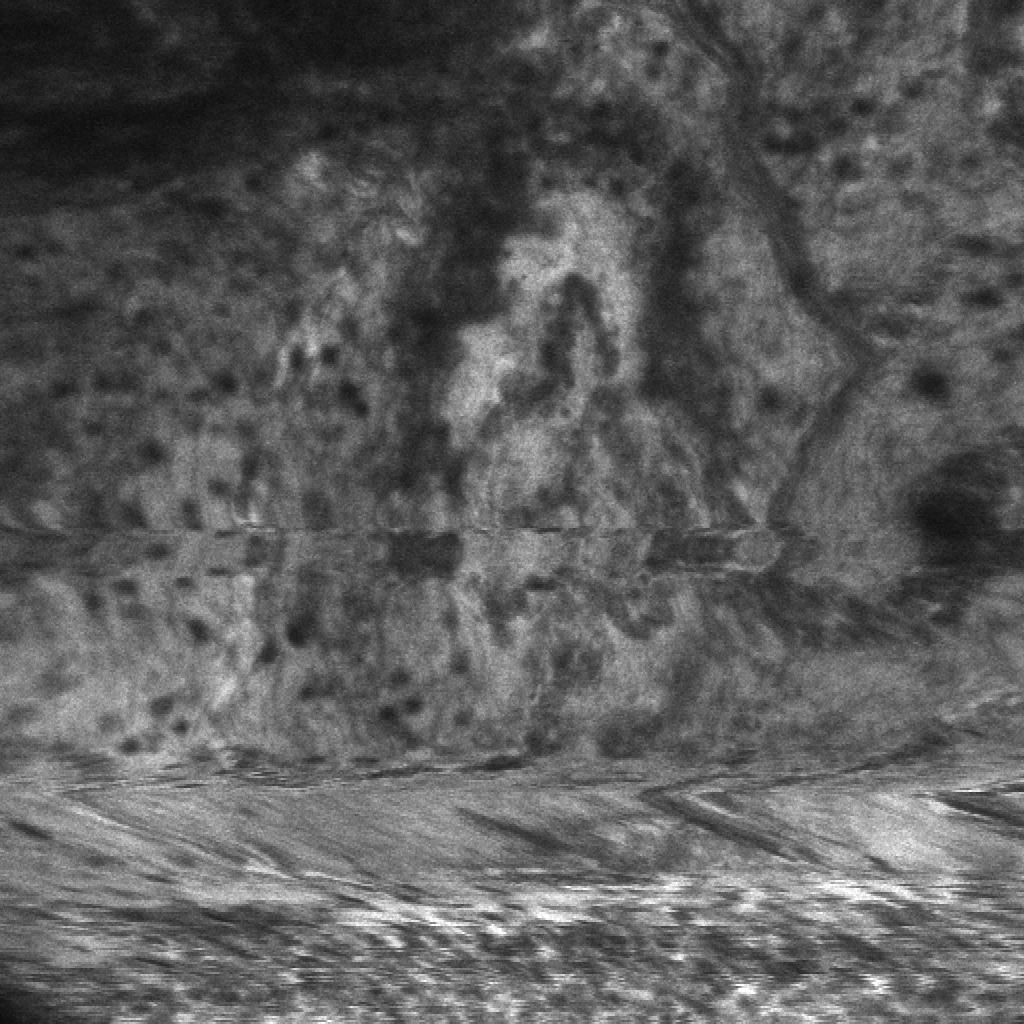}}
\caption{Nondiagnostic CLE images. Nondiagnostic images were detected through subjective evaluation done by experts. (b,e) show motion artifact and (h) shows motion artifact. Other images lack the sufficient histopathological information required for a confident diagnosis.}
\label{nondiagnostic}
\end{figure}

\begin{figure}[htb]
\centering
\subfloat[]{\includegraphics[width = 1.35in]{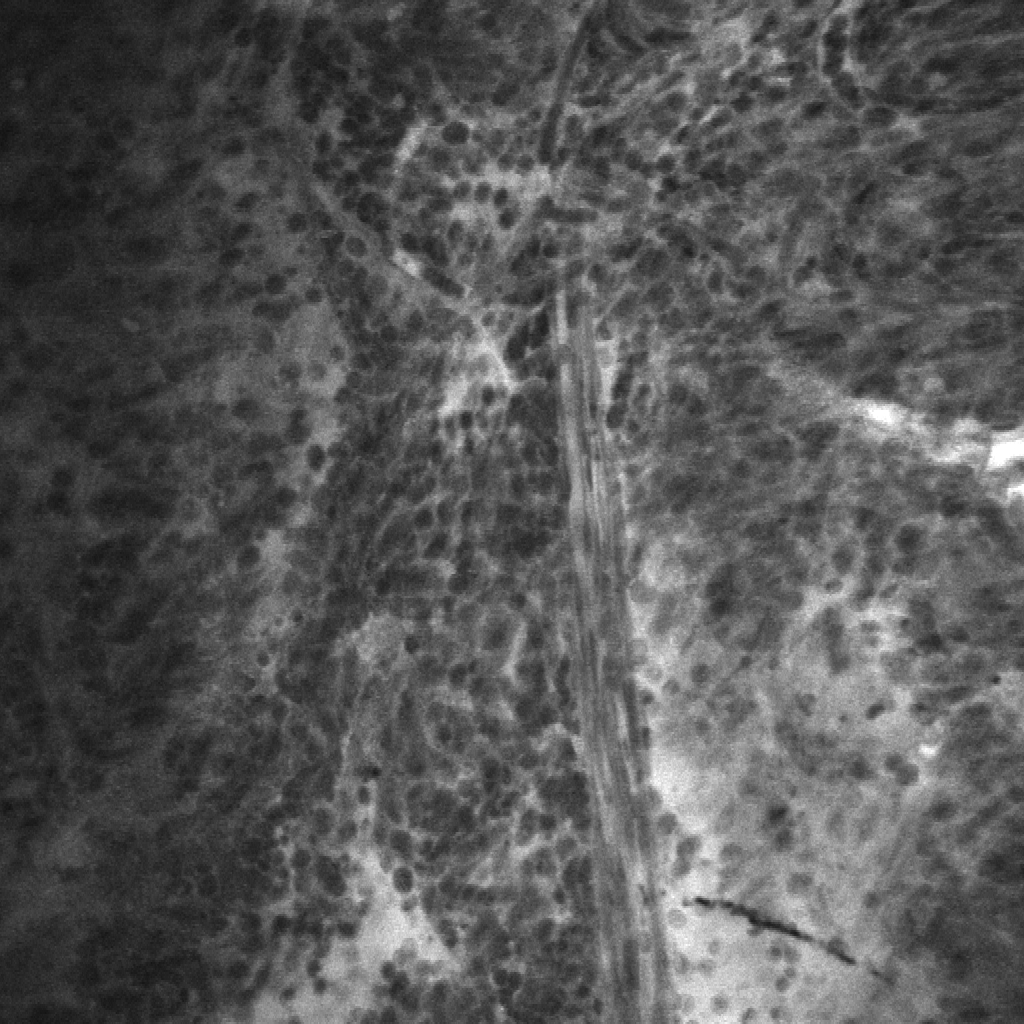}}
\subfloat[]{\includegraphics[width = 1.35in]{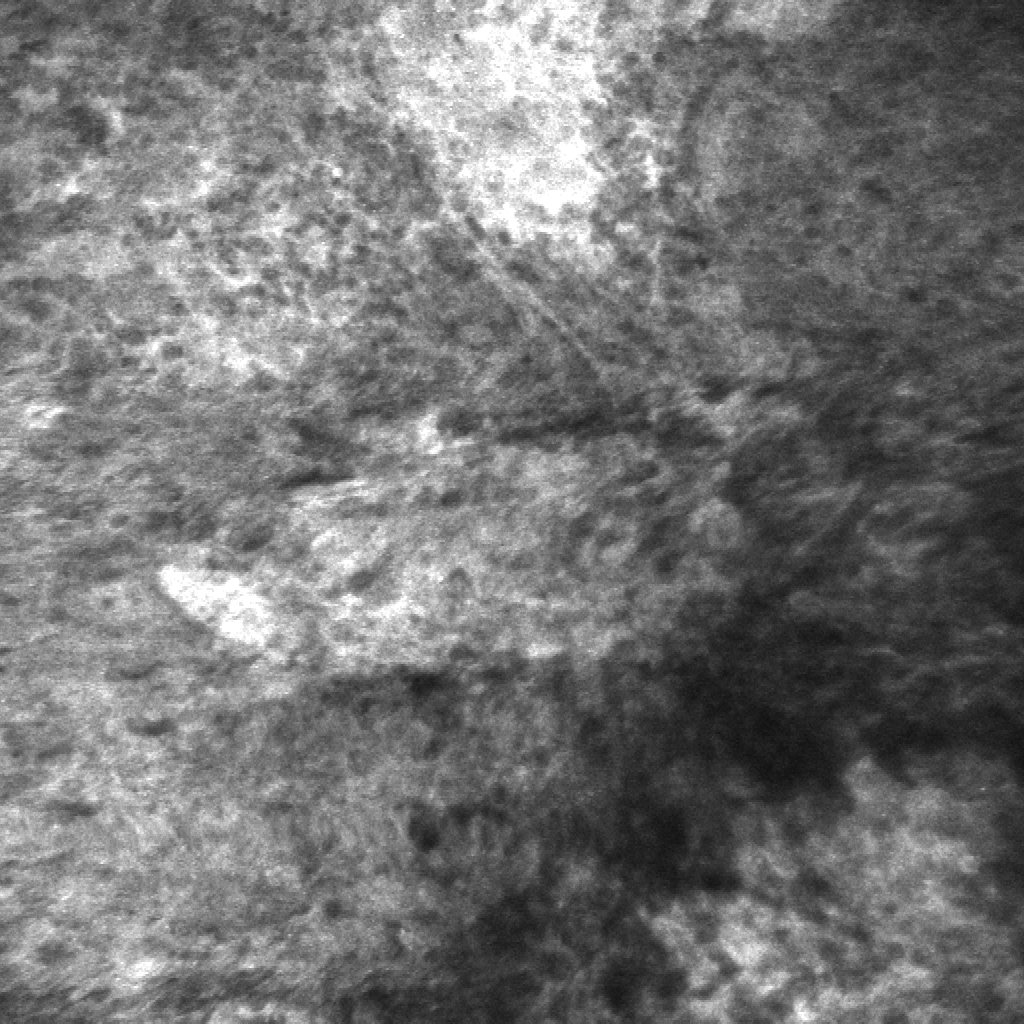}}
\subfloat[]{\includegraphics[width = 1.35in]{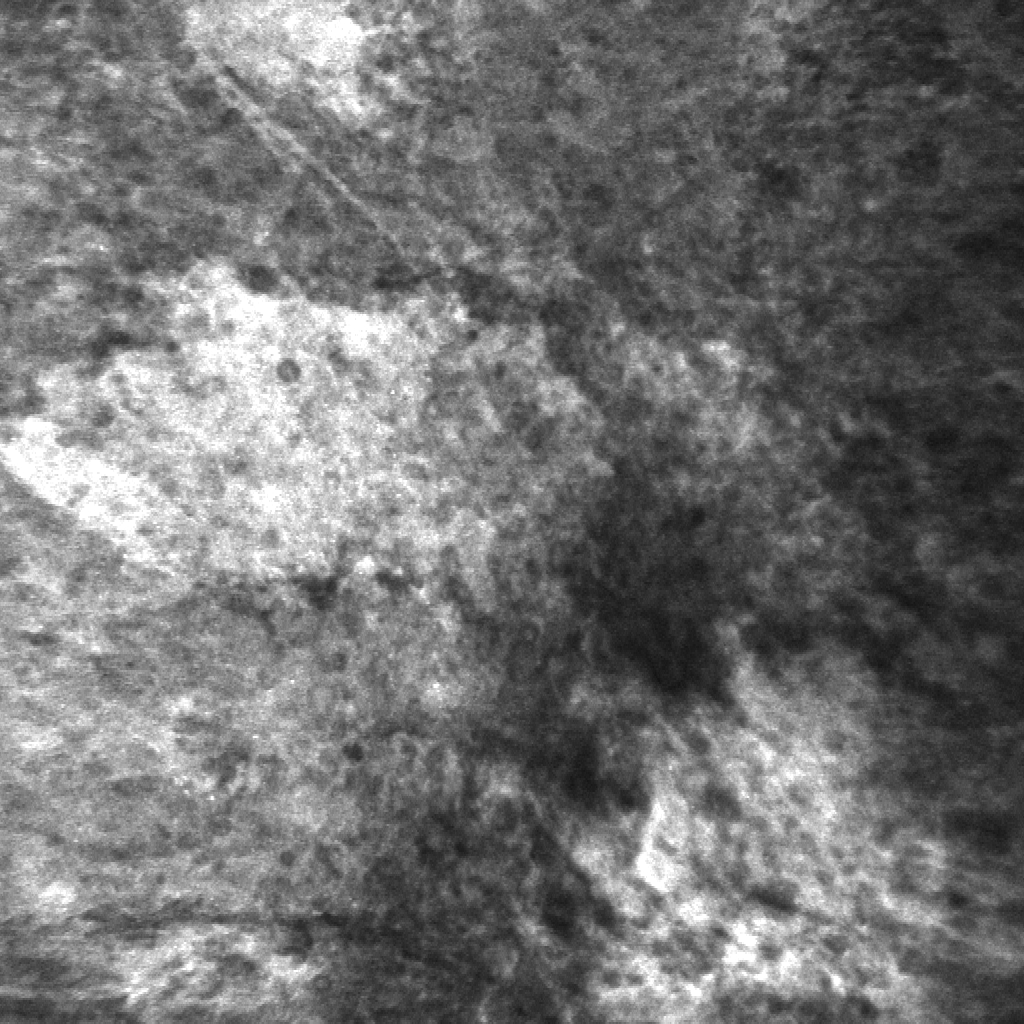}}
\subfloat[]{\includegraphics[width = 1.35in]{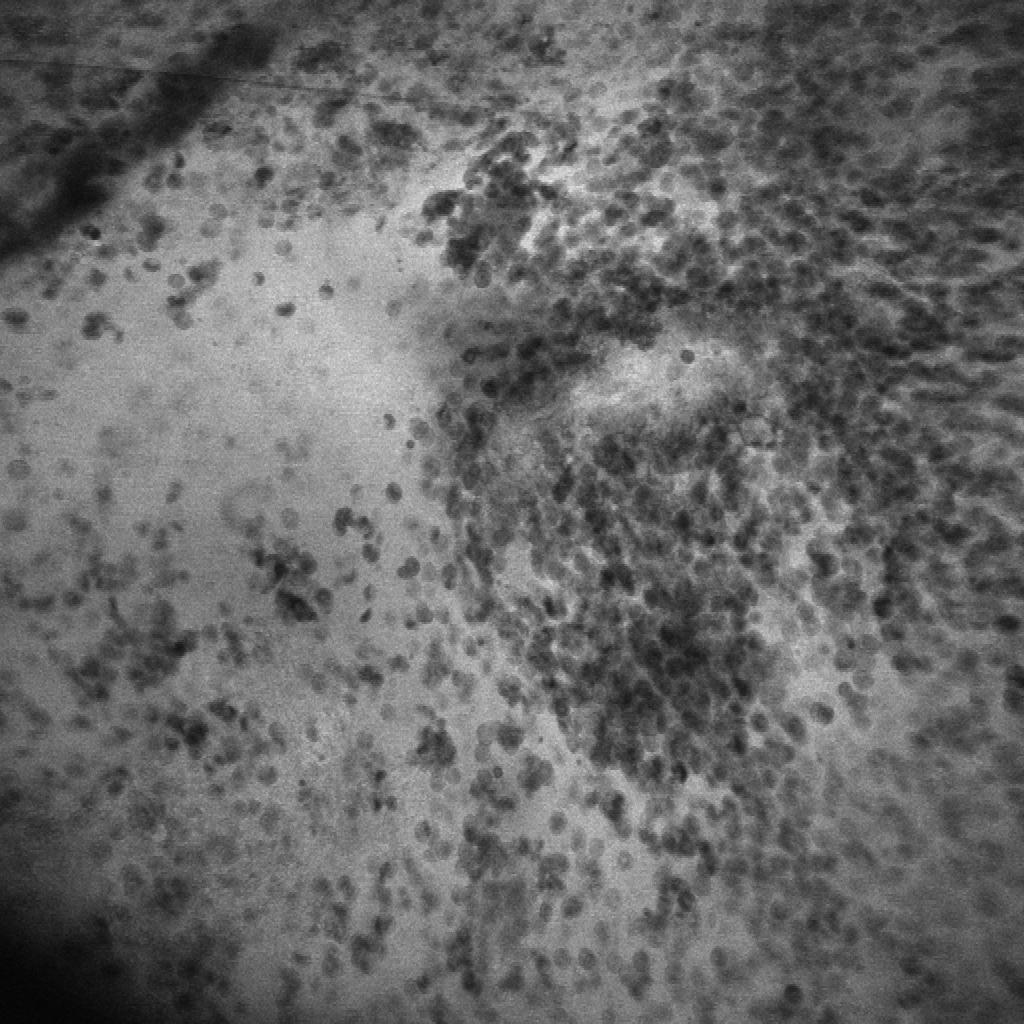}}\
\subfloat[]{\includegraphics[width = 1.35in]{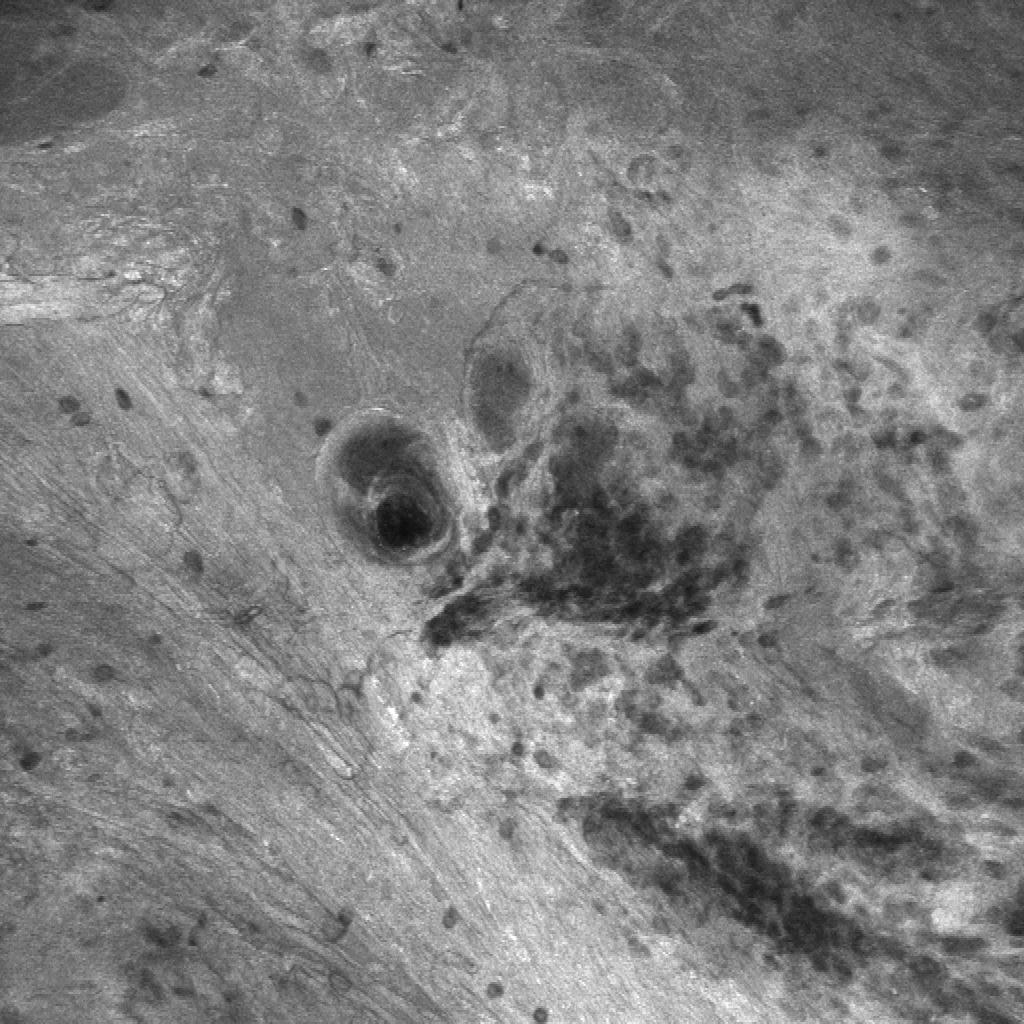}}
\subfloat[]{\includegraphics[width = 1.35in]{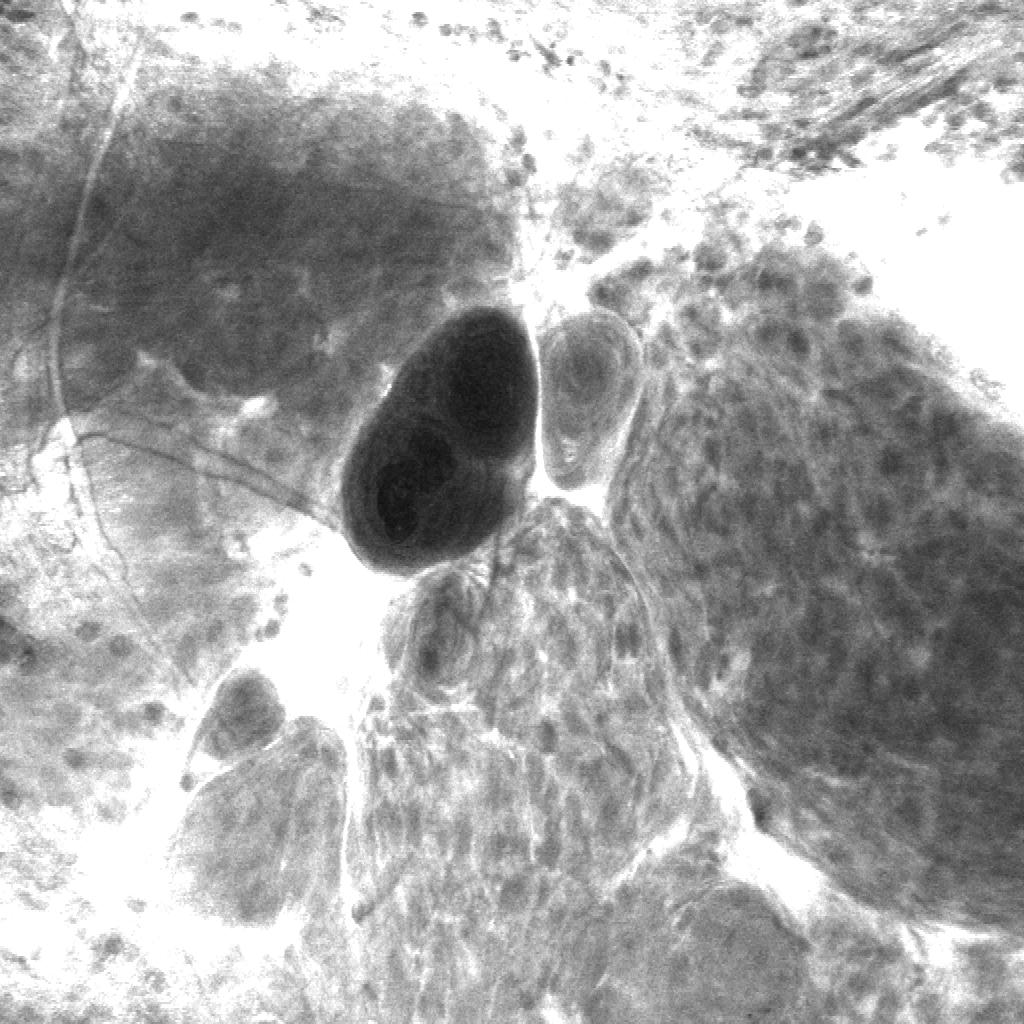}} 
\subfloat[]{\includegraphics[width = 1.35in]{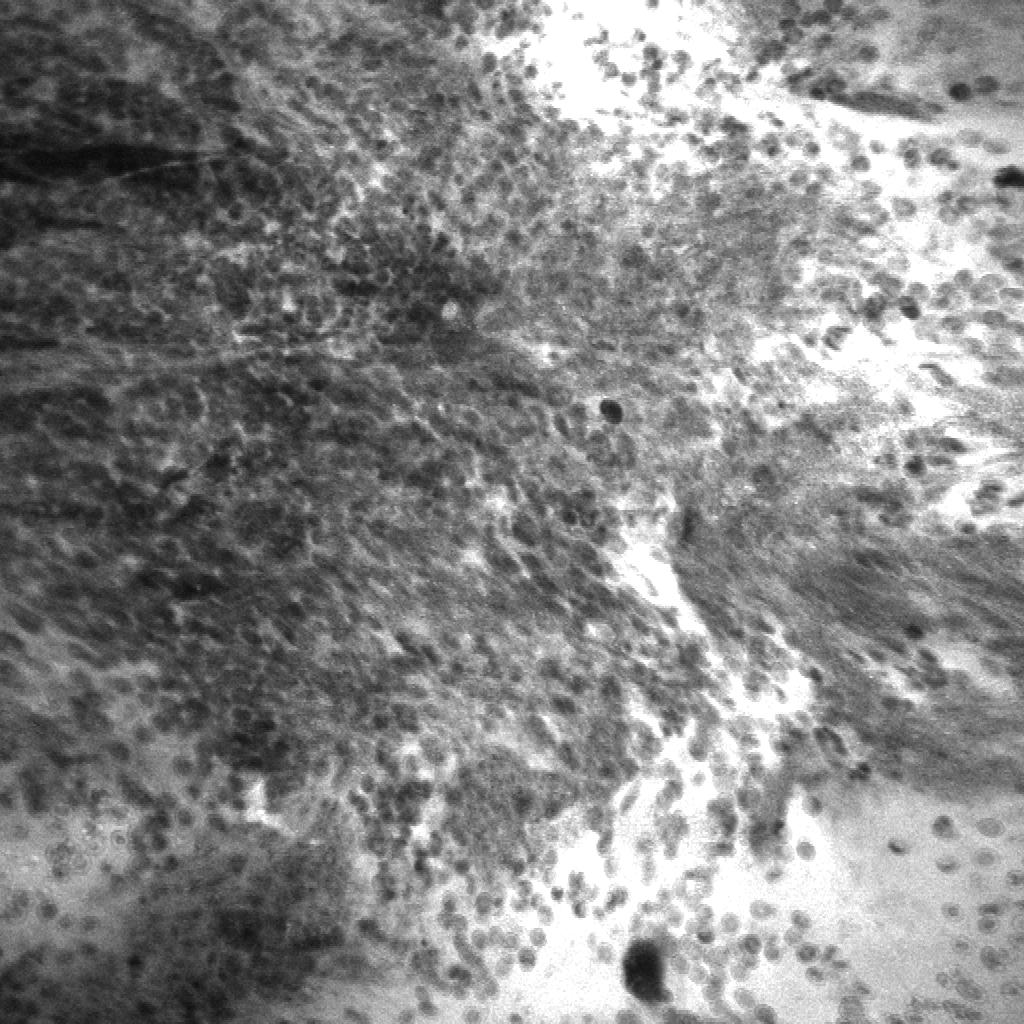}}
\subfloat[]{\includegraphics[width = 1.35in]{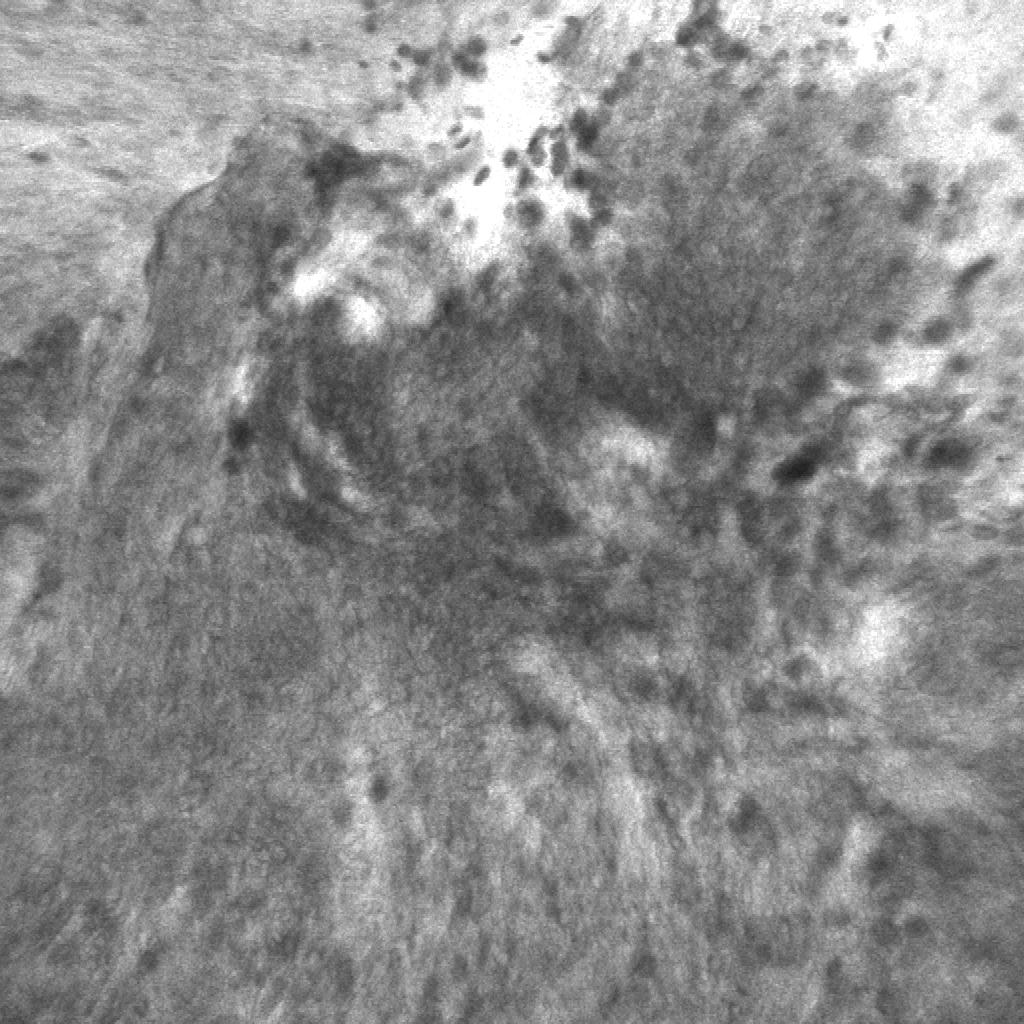}}
\caption{Diagnostic images detected by experts through subjective assessment}
\label{diagnostic}
\end{figure}

A wide range of fluorophores are able to be used for CLE in gastroenterology, but fluorophore options are limited for in vivo human brain use\cite{belykh2016intraoperative,foersch2012confocal}.  In addition, motion and blood artifacts that are present in many of the images acquired with CLE using fluorescein sodium are a barrier for delivering the instrument’s potential benefits to the surgeon. It takes time for the surgeon or pathologist to find and exclude the nondiagnostic frames and focus on diagnostic ones during the operation to make the intraoperative diagnosis \cite{martirosyan2016prospective}. In a previous study\cite{martirosyan2016prospective} our lab showed that about half of the images acquired were nondiagnostic due to abundance of motion and blood artifacts or lack of histopathological features. Figures \ref{nondiagnostic} and \ref{diagnostic} show some examples of the diagnostic and nondiagnostic CLE images from our database.\\

In our study we tried to classify the CLE images acquired from different brain tumors during surgery, using convolutional neural networks. First, we used subjective assessment to find the diagnostic CLE brain tumor images. Then, we used convolutional neural networks for making our models. We divided our data into training, validation and test sections to first train and validate the model and then test its performance on test images. We trained different CNN architectures and showed which one has better performance and
is potentially a better candidate for this task. Finally, we compared our
model performance with the results achieved from conventionally used entropy-based method. 

\section{Methods}
At this part we'll first explain a little about CLE and how the images were acquired. Then, we'll cover the mathematical methods for the CNNs used in our work.
\subsection{Image Acquisition}

The CLE (Optiscan 5.1, Optiscan Pty, Ltd.) image acquisition  system consists of a handheld miniaturized optical laser scanner designed as a rigid
probe with a 6.3 \si{\milli\meter} outer diameter and a working length
of 150 \si{\milli\meter}. A 488 \si{\nano\meter} diode laser provided incident excitation light, and fluorescent emission was detected at ~505-585 \si{\nano\meter} using a band-pass filter, via a single optical fiber
acting as both the excitation pinhole and the detection pinhole for confocal isolation of the focal plane. The detector signal was digitized synchronously with the scanning
to construct images parallel to the tissue surface (en face
optical sections).

Laser power was typically set to 550-900 \si{\micro\watt} at brain tissue; maximum power was limited to 1000 \si{\micro\watt}. A field of view of $475 \times 475$ \si{\micro\meter} (approximately
$1000\times$ magnification on a 21-inch screen) was scanned
either at $1024 \times 512$ pixels (0.8/second frame rate) or at
$1024 \times 1024$ pixels (1.2/second frame rate), with a lateral
resolution of 0.7 \si{\micro\meter} and an axial resolution (i.e., effective
optical slice thickness) of approximately 4.5 \si{\micro\meter}.

The resulting images were stored digitally and could be recorded as a time-lapse series producing essentially a digital film loop. During the procedure, a foot pedal provided remote
control of the variable confocal imaging plane depth from
the surface to a depth of 0-500 \si{\micro\meter}.
In vivo images were acquired intraoperatively during the removal of the brain tumor 5 minutes after intravenous injection of 5 \si{\milli\liter} 10\% 
FNa. 

The majority of images were obtained by using the CLE probe affixed to a Greenberg retractor arm. The retractor was tightened to a degree
that allowed both smooth movement and steady operation.
The probe was moved gently, without losing contact, along
the surface of the tissue to obtain images from several biopsy locations. Co-registration of the probe with the image guided
surgical system permitted precise intraoperative localization of the CLE imaging with the site of the biopsy.
Locations of imaging included normal brain and regions of obvious tumor, in addition to the transitional zone
between what appeared to be normal brain and tumor. Images were acquired from each biopsy location. Intraoperative CLE was performed by 4 neurosurgeons.\\
\subsubsection{Subjective Diagnostic Value Assessment}
For in vivo imaging, multiple locations within the resection bed were imaged with CLE. Tissue samples (approximately 0.5 \si{\cubic\centi\meter}) were harvested from each patient
during the procedure. For ex vivo imaging purposes, tissue samples suspicious for tumor were harvested from the
surgical field and imaged on a separate work station away
from the patient, but within the operating room. No additional fluorophore beyond intravenous FNa was used for
ex vivo imaging. Tissue samples were placed on gauze and
imaged using the probe in a freehand fashion. Multiple
images were obtained from each biopsy location.
Areas that were imaged using CLE were marked with
tissue ink so that precise locations could be validated with
conventional histology. The tissue was placed in a cassette
for standard formalin fixation and paraffin embedding.
Histological assessment was performed using standard
light microscopic evaluation of 10-\si{\micro\meter}-thick hematoxylin
and eosin (H \& E)-stained sections.

CLE images were compared with both frozen and permanent histological
sections.
Images were reviewed by a neuropathologist and 2 neurosurgeons who were not involved in the surgeries. For
each case, they analyzed the histopathological features of
corresponding CLE images and H \& E-stained frozen
and permanent sections. Images were classified as \textit{diagnostic} (i.e., the confocal images revealed identifiable histological features) or as \textit{nondiagnostic} (i.e., the image provided no identifiable histological features due to distortion by blood artifact [erythrocytes] or motion artifact).

\subsection{Convolutional Neural Network (CNN)}
\label{sec:CNN}
Convolutional Neural Network  (CNN) is a multilayer learning framework, which may consist of an input layer, a few convolutional layers and an output layer. The goal of CNN is to learn a hierarchy of feature representations. Response maps in each layer are convolved with a number of filters and further down-sampled by pooling operations.
These pooling operations aggregate values in a smaller region by down-sampling functions including max, min, and average sampling.
 In this work we adopt the softmax loss function which is given by:
\begin{equation}
L(t,y) = -\frac{1}{N}\sum_{n=1}^N \sum_{k=1}^C t_k^n log (\frac{e^{y_k^n}}{\sum_{m=1}^C e^{y_m^n}})
\end{equation}
where $t_k^n$ is the $n$-th training example's $k$-th ground truth output, and $y_k^n$ is the value of the $k$-th output layer unit in response to the $n$-th input training sample. $N$ is the number of training samples, and since we consider $2$ brain tumor type categories, $C=2$.  The learning in  CNN is based on Stochastic Gradient Descent (SGD), which includes two main operations: Forward and Back Propagation. The learning rate is dynamically lowered as training progresses. Please refer to \cite{LeCun:1998:CNI:303568.303704} for details.
%{\color{blue} So what should we put for the AlexNet and GoogLeNet? Do we need to describe details for their architecture?} {\color{red} YZ: please describe theirs architectures also, but briefly. They are basically two templates of CNN. }

The CNN architectures we used are AlexNet and GoogLeNet. We used publicly available AlexNet and GoogLeNet architectures implementation using Caffe \cite{jia2014caffe} with NVIDIA DIGITS to conduct our experiments.
\subsubsection{AlexNet} Our first CNN  has 5 convolutional layers. It starts with the input layer which is the resized version of the original image (original image size is 1024x1024 and was changed to 256x256 to fit AlexNet). Two pair of convolutional and pooling layers come in the following described layers. In each convolutional layer multiple kernels are convolved with different areas of previous layer output (receptive field) with the result progressing through a nonlinear activation function and normalization (a rectified linear unit (RLU)) to create the output of that layer.

The purpose of convolutional layers is to extract as many features as possible, while minimizing parameter numbers, partially by using the same kernel over the entire image for each following plane. Output from each convolutional layer is then fed to the next pooling layer, which replaces the output of each location in the previous plane with a summary of the surrounding pixels (in AlexNet we use maximum pooling). Pooling layers help the network to be invariant to small translations.

After two convolution-pooling combinations, the output of the last pooling layer is fed to the third convolution layer, which is followed by 
two other convolution layers (layers 6-8) and one final pooling layer (layer 9). The output of 9th layer is fed to a fully connected layer which then feeds 4096 neurons of the next fully connected layer. The process ends with the output layer, which gives the ultimate result of classification. For more details please refer to the original paper \cite{krizhevsky2012imagenet}.
\subsubsection{GoogLeNet} 
\begin{figure}[htb]
\centering
\includegraphics[width = 3in]{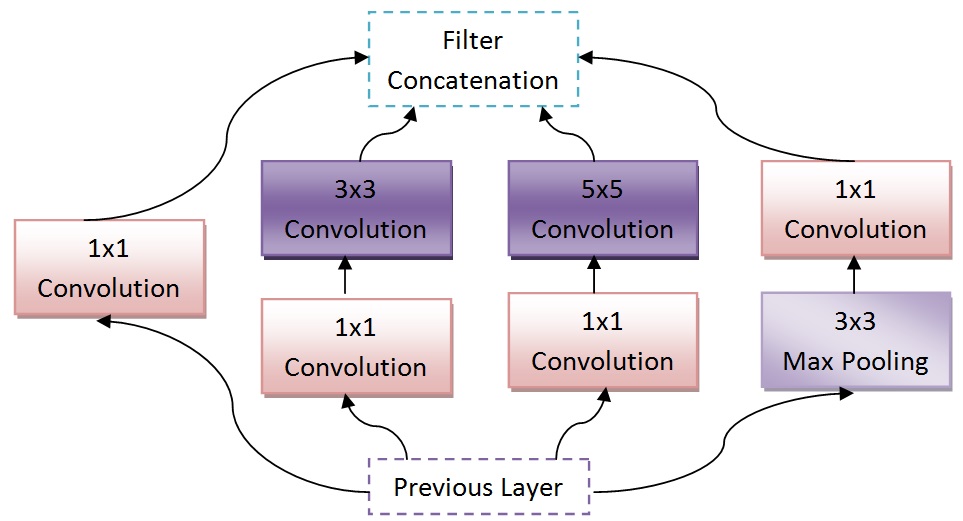}
\caption{Inception module}
\label{inception}
\end{figure}

Our second CNN has 22 layers with parameters and 9 inception modules. Each inception module is a combination of filters of size $1\times1$, $3\times3$, $5\times5$ and a  $3\times3$ max pooling, put together in parallel and the output filter banks concatenated into a single vector as the input for next stage (Fig. \ref{inception}). The network we used has similar architecture as GoogLeNet used in previous study\cite{szegedy2015going}.

\section{Experiments}
We used 2 netwroks: AlexNet and GoogLeNet. We performed a 4-fold cross validation on our data for each network. In each experiment (fold), 25 \% of images were first set apart as test images for evaluation of model. One fourth of the remaining 75 \% of images were also set apart for validation of model during training and the remaining ( 56.25 \%) were used to train the model (Table \ref{data size} ). To avoid overfitting, the training process was stopped after validation accuracy failed to further increase or loss on validation images was increasing. The trained model was then used on test images separately to evaluate the model accuracy, specificity and sensitivity. NVIDIA GeForce GTX 980 TI (6GB) was used for training and testing our networks.
\subsection{Evaluation Metrics.}
To evaluate our classifier performance we used 4 common evaluation metrics: \textit{accuracy, sensitivity, specificity and area under receiver operating characteristics (ROC) curve (AUC)}. In our study, we assumed the state of being a diagnostic image as positive and being nondiagnostic as negative. This could have been done the opposite way since it wouldn’t change the results except for producing the opposite values for sensitivity and specificity. This way, \textit{sensitivity} determines the model ability in detecting diagnostic images  and is also called true positive rate (TPR). \textit{Specificity} determines the model ability in detecting nondiagnostic images. \textit{Accuracy}, determines the model ability at true detection of diagnostic and nondiagnostic together. \cite{metz1978basic}

Each ROC curve shows true positive rate (TPR) versus false positive rate (FPR) or equivalently, sensitivity versus 1 - specificity, for different thresholds of the classifier output. \cite{fawcett2004roc} In order to use a scalar value representing the classifier performance, the area under ROC curve (AUC) is commonly used. The AUC of a
classifier is equivalent to the probability that the classifier will rank
a randomly chosen positive instance higher than a randomly chosen
negative instance. \cite{fawcett2004roc}

\begin{table}[htb]
\centering
\caption{Data used in our experiments: Number of diagnostic (Diag) and nondiagnostic (Non-Diag) images for each of four experiments in training, validation and testing stages.
}
\label{data size}
\begin{tabular}{|l|l|l|l|l|l|l|}
\hline
\multicolumn{1}{|c|}{Phase}      & \multicolumn{2}{c|}{Train}                               & \multicolumn{2}{c|}{Validation}                          & \multicolumn{2}{c|}{Test}                                \\ \hline
\multicolumn{1}{|c|}{Experiment} & \multicolumn{1}{c|}{Diag} & \multicolumn{1}{c|}{Nondiag} & \multicolumn{1}{c|}{Diag} & \multicolumn{1}{c|}{Nondiag} & \multicolumn{1}{c|}{Diag} & \multicolumn{1}{c|}{Nondiag} \\ \hline
\multicolumn{1}{|c|}{Fold 1}     & \multicolumn{1}{c|}{4626} & \multicolumn{1}{c|}{4822}    & \multicolumn{1}{c|}{1542} & \multicolumn{1}{c|}{1607}    & \multicolumn{1}{c|}{2055} & \multicolumn{1}{c|}{2143}    \\ \hline
\multicolumn{1}{|c|}{Fold 2}                           & \multicolumn{1}{c|}{4625} & \multicolumn{1}{c|}{4822} & \multicolumn{1}{c|}{1542} &           \multicolumn{1}{c|}{1607} & \multicolumn{1}{c|}{2056} & \multicolumn{1}{c|}{2143} \\ \hline
\multicolumn{1}{|c|}{Fold 3}                           & \multicolumn{1}{c|}{4625} & \multicolumn{1}{c|}{4822} & \multicolumn{1}{c|}{1542} &           \multicolumn{1}{c|}{1607} & \multicolumn{1}{c|}{2056} & \multicolumn{1}{c|}{2143} \\ \hline
\multicolumn{1}{|c|}{Fold 4}                           & \multicolumn{1}{c|}{4625} & \multicolumn{1}{c|}{4822} & \multicolumn{1}{c|}{1542} &           \multicolumn{1}{c|}{1607} & \multicolumn{1}{c|}{2056} & \multicolumn{1}{c|}{2143} \\ \hline
\end{tabular}
\end{table}
\subsection{Diagnostic Value Assessment with AlexNet}
Training the network required about 2 hours for each fold and prediction time on our test images (4199 images) was 44s (95 images/second). Results for each experiment are shown in table \ref{resultsA}. On average, we produced 90.79 \% accuracy, 90.71 \% sensitivity and 90.86 \% specificity on test images. 

In order to evaluate the reliability of our model, receiver operating characteristic (ROC) analysis was performed on the results from each experiment and area under ROC curve (AUC) was calculated (table \ref{resultsA}). Figure \ref{ROCsA:1} shows the ROC curve obtained from each fold of this experiment.  The model prediction for each image, probability of  being diagnostic or nondiagnostic and the ground truth from subjective assessment was used to perform ROC analysis in MATLAB. The same process was done for all the subsequent experiments when doing ROC analysis. The average AUC was 0.9583 in this experiment.

\begin{table}[htb]
\centering
\caption{AlexNet performance results}
\label{resultsA}
\begin{tabular}{ccccc}
\hline
Exp (\#)      & Accuracy (\%)  & Sensitivity (\%) & Specificity (\%) & AUC             \\ \hline
1             & 91.35          & 90.8             & 91.88            & 0.9607          \\
2             & 90.69          & 91.25            & 90.15            & 0.9583          \\
3             & 90.66          & 90.76            & 90.57            & 0.9584          \\
4             & 90.45          & 90.03            & 90.85            & 0.9556          \\ 
\textbf{Mean} & \textbf{90.79} & \textbf{90.71}   & \textbf{90.86}   & \textbf{0.9583}
\end{tabular}
\end{table}

\subsection{Diagnostic Value Assessment with GoogLeNet}
\begin{figure}[htb]
\centering
\subfloat[]{\includegraphics[width = 2.85in]{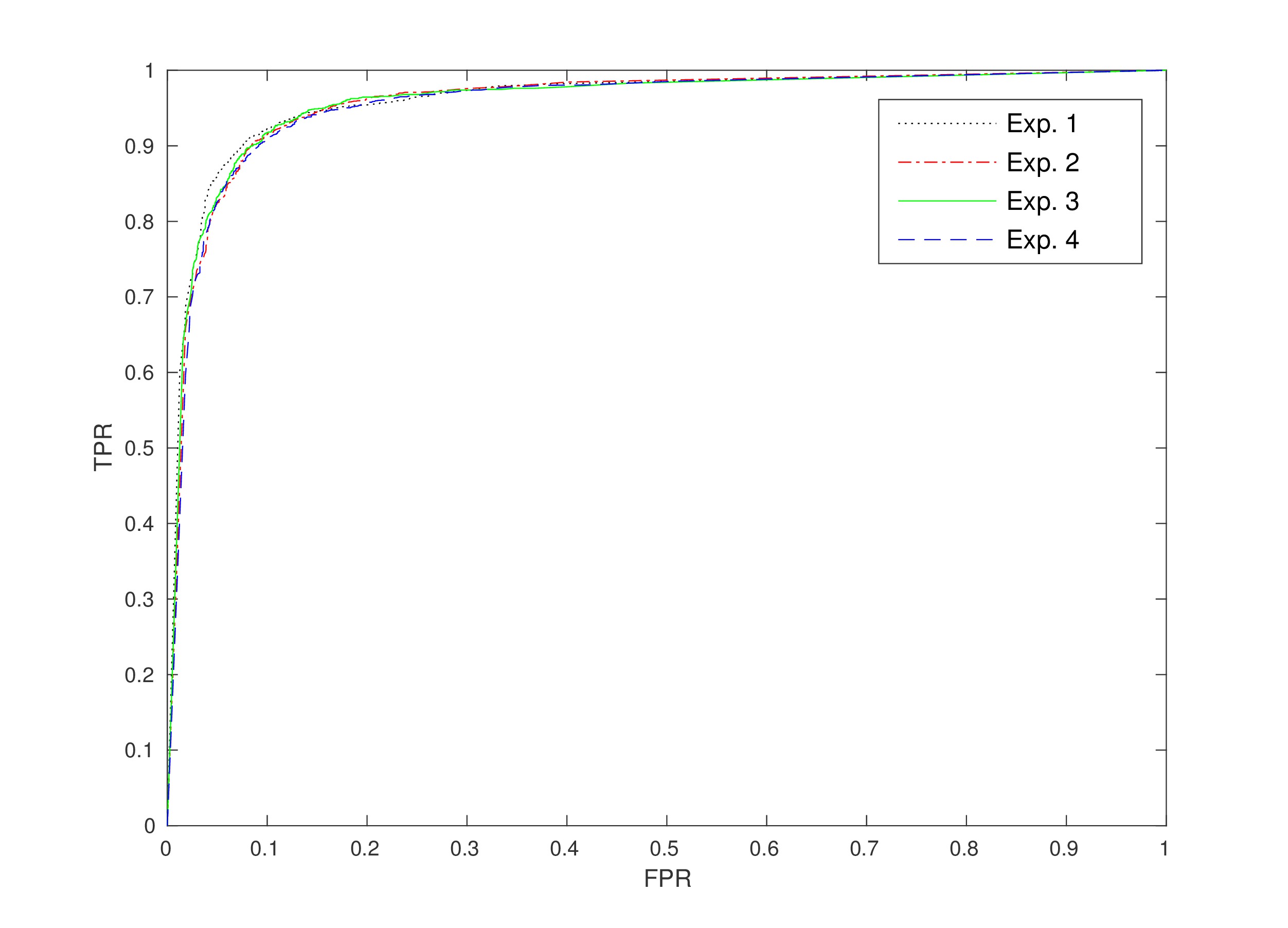}
\label{ROCsA:1}
}
\subfloat[]{\includegraphics[width = 2.85in]{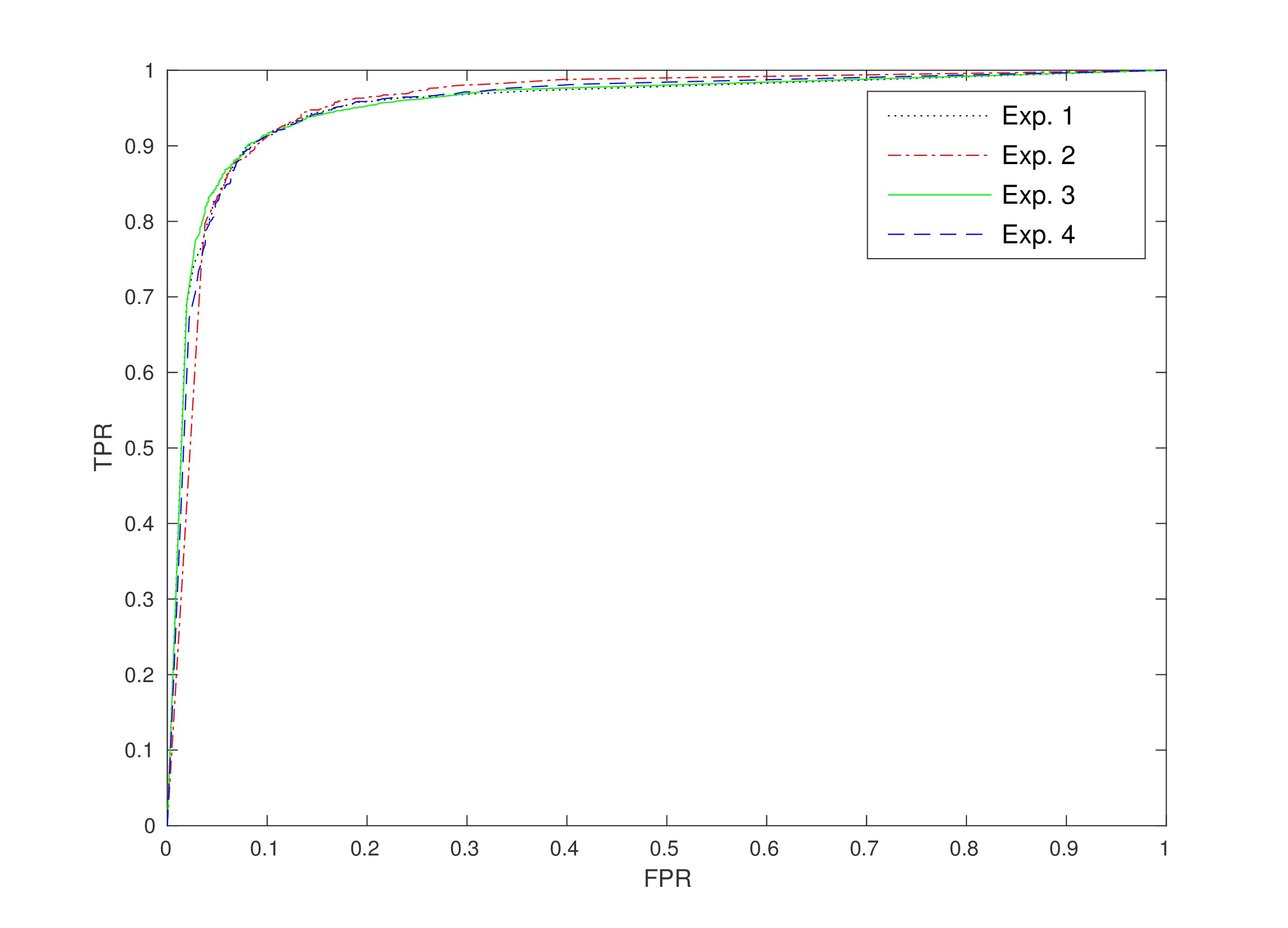}
\label{ROCsA:2}
}
\caption{ROC curves from AlexNet(a) and GoogLeNet(b) experiments. Each curve shows the sensitivity (TPR) of model vs. 1 - specificity (FPR), for one fold of the cross validation.}
\label{ROCsA}
\end{figure}

Training the network required about 9 hours for each fold and prediction time on our test images (4199 images) was 50s (84 images/second). Results for each experiment are shown in table \ref{resultsG}. On average, we produced 90.74 \% accuracy, 90.80 \% sensitivity and 90.67 \% specificity on test images.  Figure \ref{ROCsA:2} shows the mean ROC curve obtained from each fold of this experiment. The average AUC was 0.9553 in this experiment.

\begin{table}[htb]
\centering
\caption{GoogLeNet performance results}
\label{resultsG}
\begin{tabular}{ccccc}
\hline
Exp (\#)      & Accuracy (\%)  & Sensitivity (\%) & Specificity (\%) & AUC             \\ \hline
1             & 90.79          & 92.11            & 89.78            & 0.9545          \\
2             & 90.45          & 88.33            & 92.66            & 0.9561          \\
3             & 90.78          & 92.16            & 89.35            & 0.9556          \\
4             & 90.76          & 90.62            & 90.90            & 0.9551          \\ 
\textbf{Mean} & \textbf{90.74} & \textbf{90.80}   & \textbf{90.67}   & \textbf{0.9553}
\end{tabular}
\end{table}

\subsection{Diagnostic Value Assessment Using Image Entropy}
We also used entropy as a conventional method to compare the classification performance of our CNN-based model with entropy-based model. Entropy is commonly used as a measure of the information content of in an image\cite{10}. By comparing the entropy-based model predictions with the subjective diagnostic evaluation we were able to assess the validity of this assumption in CLE images. 

The entropy of all images were calculated and normalized between 0 and 1 using Matlab. We can think of the normalized entropy of an image as the probability of being informative (Generally, the image with higher entropy is more probable to be informative than the image with lower entropy). 

\begin{figure}[htb]
\centering
\subfloat[]{\includegraphics[width = 5in]{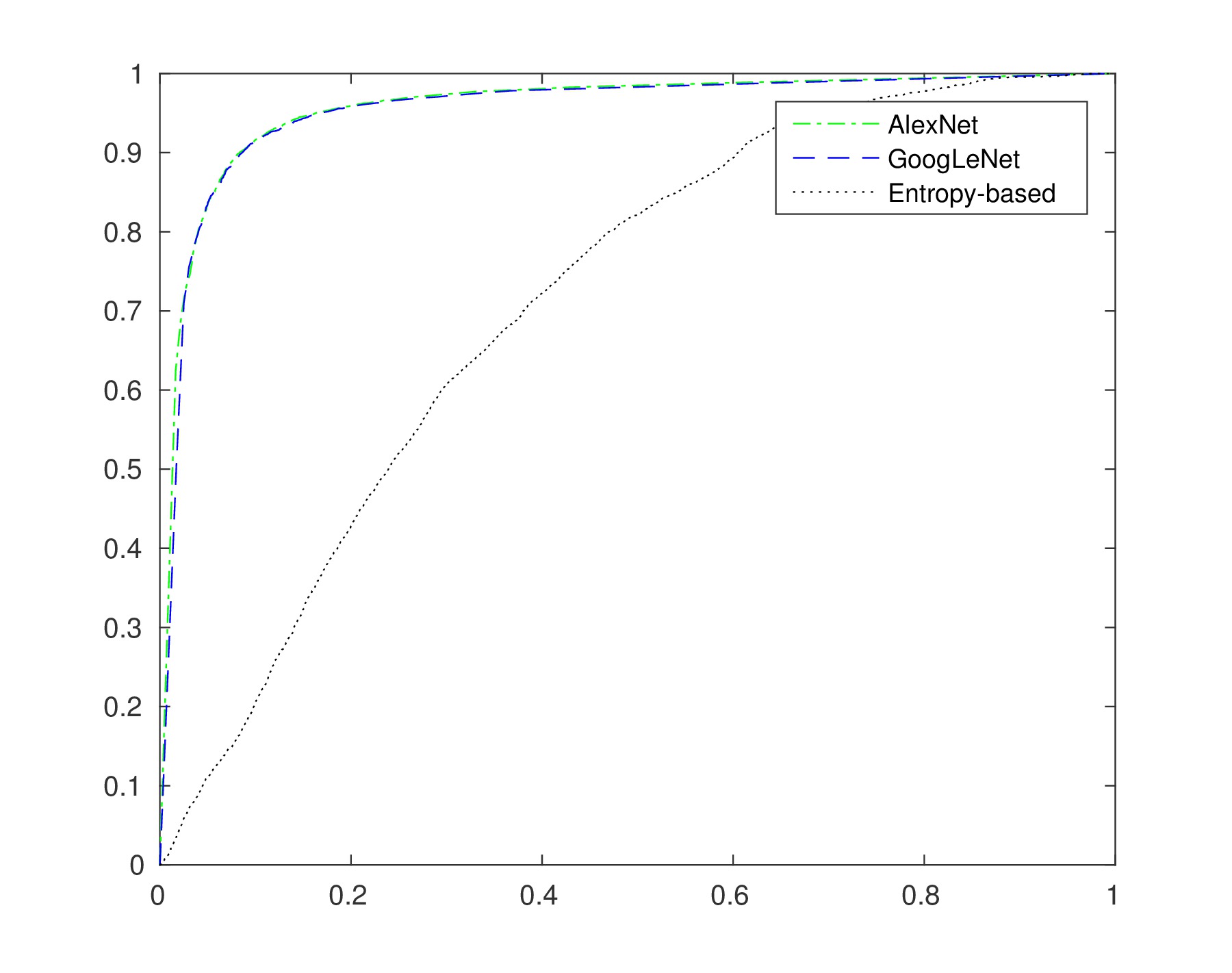}}

\caption{Average ROC curves from AlexNet, GoogLeNet and entropy-based quality assessment}
\label{ROC_all}
\end{figure}

The model prediction for each image, probability of being informative and the ground truth from subjective assessment was used to perform ROC analysis in MATLAB. Table \ref{results_all} shows the model performance. Figure \ref{ROC_all} also shows the average ROC curve (red dotted line) achieved from this experiment.  

\begin{table}[htb]
\centering
\caption{AlexNet, GoogLeNet and entropy-based performance}
\label{results_all}
\begin{tabular}{ccccc}
\hline
Model          & Accuracy (\%)  & Sensitivity (\%) & Specificity (\%) & AUC             \\ \hline
AlexNet            & \textbf{90.79} & {90.71}   & \textbf{90.86}   & \textbf{0.9583}          \\
{GoogLeNet} & {90.74} & {90.80}   & {90.67}   & {0.9553} \\
AlexNet II            & {75.95} & \textbf{98.42}   & {54.40}   & \textbf{0.9583}          \\
{GoogLeNet II} & {79.75} & {97.91}   & {62.33}   & {0.9553} \\

Entropy-based      & 57.20          & 98.20            & 17.87            & 0.7122         
\end{tabular}
\end{table}

\section{Discussion and Conclusion}
With the ongoing growth of medical imaging technologies and numerous amount of images produced, assessment of image quality is becoming more important.  During image acquisition process, multiple artifacts may be introduced to the images. Some of the most common artifacts in confocal laser endomicroscopy as a newly developed technology are blurring, noise and low/inhomogeneous contrast \cite{martirosyan2016prospective, charalampaki2015confocal}.

Blurring can be arised from maladjusted focal plane (focal blur) or relative motion between the probe and brain tissue under examination (motion blur). Environmental noise might also be introduced in the detectors. Aliasing is the origin of a variety of artifacts including unwanted jagged edges, geometric distortions and inhomogeneity of contrast \cite{hemami2010no}.

Image quality assessment (IQA) methods are usually divided into two main categories: subjective assessment  and objective assessment. A combination of these two is sometimes called hybrid image assessment. In another categorization, depending on whether the assessment is done based on comparison of the image in hand with an original image as the full reference or by itself without any reference image, assessment would be called full-reference IQA or no-reference IQA. If the comparison is based on some statistics from the original image, it would be called reduced-reference. The focus of this paper was on objective no-reference image diagnostic quality assessment.

Generally three stages are introduced \cite{hemami2010no}for objective no-reference image quality assessment: measurement of features, pooling these features in time and/or space and mapping the pooling analysis results to an estimation of the perceived quality. These features could be an estimation of one specific artifact considering a given model of that degradation \cite{ciancio2011no,corchs2014no} (e.g. blur) or a distortion-generic estimation of overall quality of the image \cite{bovik2013automatic,mittal2015no}.

Objective quality assessment of medical images has been usually criticized about its ability in estimating diagnostic quality or even visual quality of the image\cite{erickson2002irreversible}. Although the critique might still be valid for more traditional simple metrics like entropy, but our results showed that a deeply trained CNN can evaluate the diagnostic value of CLE images correlating well with the subjective assessment done by experts.

Objective quality assessment of CLE images and automatic detection of diagnostic frames has been demanded in previous studies \cite{martirosyan2016prospective}. The only attempt for pruning the nondiagnostic CLE images was done by Kamen et. al \cite{10} using entropy. Although their method  has very high sensitivity, it is very low at accuracy(57\%) and specificity (17.87\%). We thought a CNN-based model would be able to make more accurate evaluation of the diagnostic value of image since its deep multilayer architecture enables extracting abstract discriminant features, both local and global, present in a diagnostic frame. 

We created two CNN-based models (AlexNet and GoogLeNet) to test CNN capacity at predicting not just the visual quality, but the diagnostic value of a CLE image. Table \ref{results_all} shows our results from these two networks with two different decision thresholds which are used to make final prediction based on the model output: probability of being a diagnostic image. The threshold for AlexNet and GoogLeNet is set to $0.5$ and threshold for AlexNet II and GoogLeNet II is $0.00001$ to achieve the highest sensitivity possible. AlexNet and AlexNet II are making the highest performance in all evaluation metrics used. GoogLeNet and GoogLeNet II have higher accuracy and specificity but slightly lower sensitivity compared to entropy-based method. Our CNN-based models are also fast enough for using in real-time and integrates smoothly with the speed and precision of CLE imaging application in precision brain tumor surgery.

Our results showed that our trained models can provide an objective evaluation of the CLE image quality which correlates well with the subjective assessment. This separates our work from previous studies that are concerned only with the image visual quality, since it's a more reliable diagnostically oriented estimation of image quality. Automatic differentiation of diagnostic images for the further analysis would be practically useful for clinician as it would save their time for image analysis. Furthermore, similar programs might be able to suggest tumor type on the fly during image acquisition to guide a neurosurgeon in making a timely decision, translating into a shorter and more precise surgery.In our future work, we'll investigate other CNN architectures to make models with higher accuracy at diagnostic quality assessment of CLE images. 

 \acknowledgments
 This work was partially funded by the Newsome Family Endowed Chair of Neurosurgery Research at the Barrow Neurological Institute held by Dr. Preul and by funds from the Barrow Neurological Foundation.

% Note: If compiling with LaTeX+dvipdf, please ensure images generated from 
% other software packages have their bounding boxes set correctly.

% References
\bibliography{report} % bibliography data in report.bib

\begin{thebibliography}{10}

\bibitem{charalampaki2015confocal}
Charalampaki, P., Javed, M., Daali, S., Heiroth, H.-J., Igressa, A., and Weber,
  F., ``Confocal laser endomicroscopy for real-time histomorphological
  diagnosis: Our clinical experience with 150 brain and spinal tumor cases,''
  {\em Neurosurgery}~{\bf 62},  171--176 (2015).

\bibitem{sanai2011intraoperative}
Sanai, N., Eschbacher, J., Hattendorf, G., Coons, S.~W., Preul, M.~C., Smith,
  K.~A., Nakaji, P., and Spetzler, R.~F., ``Intraoperative confocal microscopy
  for brain tumors: a feasibility analysis in humans,'' {\em Neurosurgery}~{\bf
  68},  ons282--ons290 (2011).

\bibitem{belykh2016intraoperative}
Belykh, E., Martirosyan, N.~L., Yagmurlu, K., Miller, E.~J., Eschbacher, J.~M.,
  Izadyyazdanabadi, M., Bardonova, L.~A., Byvaltsev, V.~A., Nakaji, P., and
  Preul, M.~C., ``Intraoperative fluorescence imaging for personalized brain
  tumor resection: Current state and future directions,'' {\em Frontiers in
  Surgery}~{\bf 3} (2016).

\bibitem{foersch2012confocal}
Foersch, S., Heimann, A., Ayyad, A., Spoden, G.~A., Florin, L., Mpoukouvalas,
  K., Kiesslich, R., Kempski, O., Goetz, M., and Charalampaki, P., ``Confocal
  laser endomicroscopy for diagnosis and histomorphologic imaging of brain
  tumors in vivo,'' {\em PLoS One}~{\bf 7}(7),  e41760 (2012).

\bibitem{martirosyan2016prospective}
Martirosyan, N.~L., Eschbacher, J.~M., Kalani, M. Y.~S., Turner, J.~D., Belykh,
  E., Spetzler, R.~F., Nakaji, P., and Preul, M.~C., ``Prospective evaluation
  of the utility of intraoperative confocal laser endomicroscopy in patients
  with brain neoplasms using fluorescein sodium: experience with 74 cases,''
  {\em Neurosurgical focus}~{\bf 40}(3),  E11 (2016).

\bibitem{LeCun:1998:CNI:303568.303704}
LeCun, Y. and Bengio, Y., ``The handbook of brain theory and neural networks,''
  ch.~Convolutional networks for images, speech, and time series,  255--258,
  MIT Press, Cambridge, MA, USA (1998).

\bibitem{jia2014caffe}
Jia, Y., Shelhamer, E., Donahue, J., Karayev, S., Long, J., Girshick, R.,
  Guadarrama, S., and Darrell, T., ``Caffe: Convolutional architecture for fast
  feature embedding,'' {\em arXiv preprint arXiv:1408.5093}  (2014).

\bibitem{krizhevsky2012imagenet}
Krizhevsky, A., Sutskever, I., and Hinton, G.~E., ``Imagenet classification
  with deep convolutional neural networks,'' in [{\em Advances in neural
  information processing systems}{\nolinebreak\hspace{0.1em}]},   1097--1105
  (2012).

\bibitem{szegedy2015going}
Szegedy, C., Liu, W., Jia, Y., Sermanet, P., Reed, S., Anguelov, D., Erhan, D.,
  Vanhoucke, V., and Rabinovich, A., ``Going deeper with convolutions,'' in
  [{\em Proceedings of the IEEE Conference on Computer Vision and Pattern
  Recognition}{\nolinebreak\hspace{0.1em}]},   1--9 (2015).

\bibitem{metz1978basic}
Metz, C.~E., ``Basic principles of roc analysis,'' in [{\em Seminars in nuclear
  medicine}{\nolinebreak\hspace{0.1em}]},   {\bf 8}(4),  283--298, Elsevier
  (1978).

\bibitem{fawcett2004roc}
Fawcett, T., ``Roc graphs: Notes and practical considerations for
  researchers,'' {\em Machine learning}~{\bf 31}(1),  1--38 (2004).

\bibitem{10}
Kamen, A., Sun, S., Wan, S., Kluckner, S., Chen, T., Gigler, A.~M., Simon, E.,
  Fleischer, M., Javed, M., Daali, S., et~al., ``Automatic tissue
  differentiation based on confocal endomicroscopic images for intraoperative
  guidance in neurosurgery,'' {\em BioMed research international}~{\bf 2016}
  (2016).

\bibitem{hemami2010no}
Hemami, S.~S. and Reibman, A.~R., ``No-reference image and video quality
  estimation: Applications and human-motivated design,'' {\em Signal
  processing: Image communication}~{\bf 25}(7),  469--481 (2010).

\bibitem{ciancio2011no}
Ciancio, A., da~Costa, A. L. N.~T., da~Silva, E.~A., Said, A., Samadani, R.,
  and Obrador, P., ``No-reference blur assessment of digital pictures based on
  multifeature classifiers,'' {\em IEEE Transactions on image processing}~{\bf
  20}(1),  64--75 (2011).

\bibitem{corchs2014no}
Corchs, S., Gasparini, F., and Schettini, R., ``No reference image quality
  classification for jpeg-distorted images,'' {\em Digital Signal
  Processing}~{\bf 30},  86--100 (2014).

\bibitem{bovik2013automatic}
Bovik, A.~C., ``Automatic prediction of perceptual image and video quality,''
  {\em Proceedings of the IEEE}~{\bf 101}(9),  2008--2024 (2013).

\bibitem{mittal2015no}
Mittal, A., Moorthy, A.~K., Bovik, A.~C., Chen, C.~W., Chatzimisios, P.,
  Dagiuklas, T., and Atzori, L., ``No-reference approaches to image and video
  quality assessment,'' {\em Multimedia Quality of Experience (QoE): Current
  Status and Future Requirements} ,  99 (2015).

\bibitem{erickson2002irreversible}
Erickson, B.~J., ``Irreversible compression of medical images,'' {\em Journal
  of Digital Imaging}~{\bf 15}(1),  5--14 (2002).

\end{thebibliography}
\bibliographystyle{spiebib} % makes bibtex use spiebib.bst

\end{document}